\documentclass[conference,a4paper]{IEEEtran}
\IEEEoverridecommandlockouts

\usepackage[hidelinks]{hyperref}
\usepackage[cmex10]{amsmath}
\usepackage{amssymb,amsfonts}
\usepackage{dblfloatfix}
\usepackage{rotating}
\usepackage[ruled,vlined]{algorithm2e}
\usepackage{graphicx}
\graphicspath{{Figures/PDF/}{Figures/PNG/}}
\usepackage{flushend}
\usepackage{booktabs}
\usepackage{siunitx}
\usepackage[numbers,compress]{natbib}
\usepackage{texnames}
\usepackage{multirow}
\usepackage{bm,bbm}
\usepackage{orcidlink}
\usepackage{subcaption}

\usepackage[capitalize]{cleveref}
\crefname{section}{Sec.}{Secs.}
\Crefname{section}{Section}{Sections}
\Crefname{table}{Table}{Tables}
\crefname{table}{Tab.}{Tabs.}

\begin{document}

\title{Feature-Space Oversampling for Addressing Class Imbalance in SAR Ship Classification
\thanks{This work was supported by National Recovery and Resilience Plan (NRRP), Mission 4 Component 2 Investment 1.4 - Call for tender No. 3138 of 16 December 2021, rectified by Decree n.3175 of 18 December 2021  of Italian Ministry of University and Research funded by the European Union – NextGenerationEU. Award Number: Project code CN\_00000033, Concession Decree No. 1034  of 17 June 2022 adopted by the Italian Ministry of University and Research,  CUP D33C22000960007, Project title “National Biodiversity Future Center - NBFC”. \\ This work was conducted during an exchange period at Cardiff University.}
}

\author{	\IEEEauthorblockN{Ch Muhammad Awais\orcidlink{0009-0001-4589-4103}}
	\IEEEauthorblockA{\textit{ISTI-CNR \& University of Pisa}\\
		Pisa, Italy\\
		chmuhammad.awais@phd.unipi.it}
	\and
	\IEEEauthorblockN{Marco Reggiannini\orcidlink{0000-0002-4872-9541}}
	\IEEEauthorblockA{\textit{ISTI-CNR \& NBFC}\\
		Pisa, Palermo, Italy\\
		marco.reggiannini@isti.cnr.it}
	\and
	\IEEEauthorblockN{Davide Moroni\orcidlink{0000-0002-5175-5126}}
	\IEEEauthorblockA{\textit{ISTI-CNR}\\
		Pisa, Italy\\
		davide.moroni@isti.cnr.it}
        \and
	\IEEEauthorblockN{Oktay Karaku\c{s}\orcidlink{0000-0001-8009-9319}}
	\IEEEauthorblockA{
    \textit{Cardiff University}\\ Cardiff, UK. \\
		karakuso@cardiff.ac.uk}
        
}

\maketitle
\begin{abstract}
SAR ship classification faces the challenge of long-tailed datasets, which complicates the classification of underrepresented classes. Oversampling methods have proven effective in addressing class imbalance in optical data. In this paper, we evaluated the effect of oversampling in the feature space for SAR ship classification. We propose two novel algorithms inspired by the Major-to-minor (M2m) method M2m$_f$, M2m$_u$. The algorithms are tested on two public datasets, OpenSARShip (6 classes) and FuSARShip (9 classes), using three state-of-the-art models as feature extractors: ViT, VGG16, and ResNet50. Additionally, we also analyzed the impact of oversampling methods on different class sizes. The results demonstrated the effectiveness of our novel methods over the original M2m and baselines, with an average F1-score increase of 8.82\% for FuSARShip and 4.44\% for OpenSARShip.
\end{abstract}


\section{Introduction}
\label{sec:introduction}
Synthetic aperture radar (SAR), in conjunction with the Automatic Identification System (AIS), plays a crucial role in maritime traffic monitoring applications, including deep learning-based SAR ship classification \cite{rs14215288}. However, SAR ship classification datasets often face two significant challenges: insufficient data and high class imbalance \cite{zhang2021imbalanced}. Deep learning models require large datasets for effective training, and when data is limited, transfer learning techniques are commonly employed \cite{hosna2022transfer}. In traditional transfer learning, the feature extraction layers of a pretrained model are frozen, and a classifier is trained on the extracted features. To address data imbalance, data augmentation techniques are typically used to enhance the dataset \cite{shorten2019survey}. However, these methods frequently fail for SAR data due to the presence of significant noise and the inherently low resolution of SAR images.

The feature space provides an alternative approach to address these challenges by generating synthetic features using techniques like oversampling in feature space \cite{gosain2017handling}. Oversampling leverages features extracted from a pretrained network to create synthetic features, thereby mitigating class imbalance. Previous works have explored feature-space methods for SAR ship classification. For example, Li et al.\cite{9812628} compared oversampling techniques for three-class classification on OpenSARShip and FuSARShip, Zhang et al. \cite{9390936} applied oversampling on six classes of OpenSARShip using accuracy as a performance metric, and Xie et al. \cite{10640754} proposed a novel loss function evaluated on six classes of FuSARShip.
\begin{figure}
    \centering
    \includegraphics[width=1\linewidth]{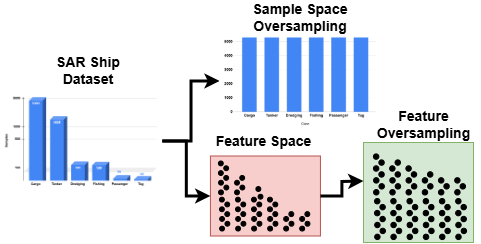}
    \caption{Sampling Overview}
    \label{fig:sampling_over}
\end{figure}

The oversampling technique Major-to-minor (M2m) \cite{kim2020m2m} is particularly effective in addressing data imbalance in sample space. M2m generates synthetic samples for minority classes by utilizing samples from majority classes, thereby increasing the representation of underrepresented classes without using the samples from minority classes. M2m has not been explored for SAR ship classification, in this study, we compare the original M2m with two modified versions of M2m that perform oversampling in feature space (\Cref{fig:sampling_over}). We adopted the F1-score \cite{10701968}, a metric well-suited for imbalanced data, to evaluate the effectiveness of the proposed framework comprehensively. Our main contributions are as follows:
\begin{itemize}
    \item We propose two effective oversampling frameworks inspired by the Major-to-minor (M2m) method, specifically designed to address the challenges of class imbalance in SAR ship classification.
    \item We improve the classification performance across datasets, highlighting the generalizability of our approach.
\end{itemize}

The rest of the paper is arranged as follows:  \Cref{sec:methodology} outlines our proposed methodology. \Cref{sec:results} presents the results of our experiments, while \Cref{sec:discussion} discusses these findings. Finally, \Cref{sec:conclusion} concludes the paper.

\section{Methodology}
\label{sec:methodology}
\begin{figure}[hbt]
    \centering
    \includegraphics[width=1.0\linewidth]{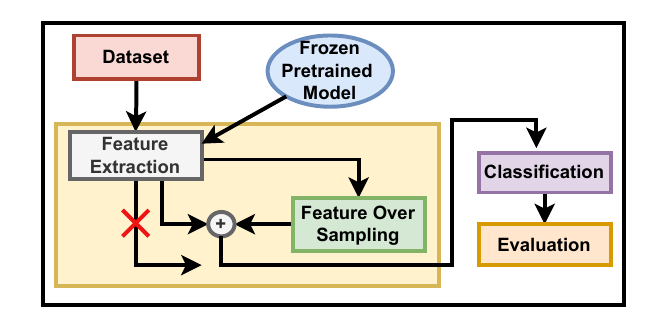}
    \caption{(1) Dataset: SAR data; (2) Feature extraction: pretrained model with frozen layers; (3) Feature oversampling: method to generate synthetic features; (4) Classification: assigns labels to the features; (5) Evaluation: compares the predicted labels with the original labels to assess performance.}
    \label{fig:method}
\end{figure}

The proposed approach consists of two oversampling methods in feature space to mitigate the imbalance issue (see \Cref{fig:method}). Traditionally, the training pipeline can be divided into three parts: (1) \textit{feature extractor}: usually a pre-trained network which takes an input and provides features. Feature extractor is often frozen, meaning that the weights are not updated while training; (2) \textit{classifier}: a model that takes the features extracted by the feature extractor and predicts the class of the extracted features; (3) \textit{loss function}: takes the predicted output of the classifier and compares it with the actual label, then updates the weights of the classifier (since the extractor is frozen). In our methodology, the features extracted by the feature extractor are exploited to help the classifier learn better; in \cref{fig:method}, it is presented as the ``Feature Over Sampling" block.

The two proposed oversampling methods, M2m$_f$ and M2m$_u$, 
are compared with the baseline and the original M2m (M2m$_{orig}$). M2m$_f$ (described in \cref{algo:Algo_m2m}), is based on the implementation provided in the original M2m paper but operates in the feature space instead of the sample space. It takes extracted features and increases the number of features representative of the minority classes by using features from the majority classes.

M2m$_f$ first identifies the minority classes with samples fewer than a defined minority threshold $M_v$, which is set by analyzing the samples of all classes. Second, centroids ($ctr_m$) are calculated from the features of minority classes (average feature values of each class). These centroids are used to generate synthetic features by introducing minor changes to the majority class features. A synthetic feature is retained only if it differs by a defined minimum distance ($d_t=0.5$) from existing synthetic features (to keep samples unique). Lastly, only ($M_v$) synthetic features are retained, and these features are appended to the original dataset ($D$). The updated features are then used to train the models.

\begin{algorithm}[hbt]
	\DontPrintSemicolon
	\SetKwInput{KwCompute}{Compute}
	\SetKwInput{KwDefine}{Define}
	\caption{M2m$_f$}\label{algo:Algo_m2m}
	\KwData{Set of extracted Features $D$, Minority Value $M_v$, Distance Threshold $d_t$, lambda $\lambda=0.1$.}
    $c_m = $ class indexes $c$ such that $len(D[c])<M_v$\\
    $ctr_m = \{\text{class centroids } ctr[c],\, c \in c_m\}$\\
    \vspace{3px}
    $D_M = \{D \setminus D_m$\} where $D_m = \bigcup\limits_{c \in c_m} D[c]$\\
    \textbf{Initialize synthetic features} $S$: $S[c]=\emptyset,\, c \in c_m$\\
    \For{$c \in c_{m}$}{
    \For{$\mathbf f \in D_M$}{
         $\mathbf{s} = \mathbf{f} + \lambda(\text{ctr}[c] - \mathbf{f})$\\
         $d_m =$ \text{min dist of $\mathbf{s}$ from any feature in $S[c]$}\\
         \If{$d_m > d_t$}{
             \textbf{Add} $\mathbf{s}$ to $S[c]$
             }
        \If{$len(S[c]) > M_v$}{
            break inner loop: Go to next minority class
            }
        }
    }
    \KwOut {$S$}
\end{algorithm}

The second variation of the feature-based algorithm, M2m$_u$, follows a similar pattern to its predecessor but introduces a different eligibility criterion for retaining synthetic samples. Instead of evaluating the distance from existing synthetic features, this variation retains a new synthetic sample only if it is deemed ``\textit{similar}" to features in the original dataset. This adjustment ensures that synthetic features remain closely aligned with existing ones. To achieve this, \textit{Cosine similarity} is calculated between the new synthetic feature, $\mathbf{s}$, and the features of the same class $D[c]$ (as in \Cref{eq:synth_eqn}). Synthetic features are retained only if their similarity exceeds a predefined threshold ($sim_t =0.8$). 

\begin{equation}
    \label{eq:synth_eqn}
    sim = cosine(s, D[c])
\end{equation}

The baseline comprised models that combined frozen feature extraction algorithms with a non-frozen classifier in a pipeline. The M2m$_{orig}$ implementation is taken from the code provided by the original paper \cite{kim2020m2m}, by simply increasing the number of samples based on the updated ratios using ``\textit{WeightedRandomSampler}'' from ``\textit{pytorch}''. 

\begin{table}[hbt]
\centering
\caption{Datasets and Minority Values $M_v$. (\small{\textbf{bold} represents the minority classes $c_m$ which were oversampled with the $M_v$ features})}
    \begin{tabular}{rrp{5.6cm}}
    \toprule
    \multicolumn{1}{l}{Dataset} & \multicolumn{1}{l}{$M_v$} & Classes \\
    \toprule
    \multicolumn{1}{r}{\multirow{3}[0]{*}{\begin{turn}{30}FuSARShip\end{turn}}} & 800   & \textbf{Fishing}, Cargo \\\cline{2-3}
          & 500   & Fishing, Cargo, \textbf{Bulk}, \textbf{Tanker} \\\cline{2-3}
          & 500   & Fishing, Cargo, \textbf{Bulk}, \textbf{Tanker}, \textbf{Tug}, \textbf{Container}, \textbf{Passenger}, \textbf{GeneralCargo}, Dredging \\\toprule
    \multicolumn{1}{r}{\multirow{3}[0]{*}{\begin{turn}{30}OpenSARShip\end{turn}}} & 3000  & \textbf{Tanker}, Cargo \\\cline{2-3}
          & 250   & Tanker, Cargo, \textbf{Fishing}, \textbf{Bulk} \\\cline{2-3}
          & 250   & Tanker, Cargo, \textbf{Fishing}, \textbf{Bulk}, \textbf{Tug}, \textbf{Passenger} \\\bottomrule
    \end{tabular}%
\label{tab:data_ratios}
\end{table}

The dataset and their corresponding minority values ($M_v$) are presented in \Cref{tab:data_ratios}. The $M_v$ values were determined by analyzing the sample distribution within the dataset. For instance, in the FuSARShip 2-class dataset, the ``Cargo" class contained approximately 1,700 samples, while the ``Fishing" class had only around 800. To address the aforementioned imbalance, an additional 800 samples were generated for the ``Fishing" class. Similarly, in the 4-class dataset, other classes contained fewer than 500 samples, so 500 additional samples were generated for each minority class, and so forth.

Three pretrained networks (VGG16 \cite{simonyan2014very}, ResNet50 \cite{he2016deep}, and ViT-16\_base224 \cite{dosovitskiy2020image}) are utilized for feature extraction. For VGG and ResNet, the images were resized down to 64$\times$64, whereas for ViT, the image size was set to 224$\times$224. The features from feature extractors undergo average pooling and flattening before being processed by a fully connected block. This block consists of a linear layer, followed by a ReLU activation function, a dropout layer for regularization, and a final linear layer to generate the class predictions.

The models were trained for 30 epochs using a batch size of 32, with a learning rate of 0.0001. The training process utilized the \textit{Adam} optimizer and \textit{CrossEntropy} as the loss function. Python libraries such as \textit{PyTorch} and \textit{timm} were employed for implementing and training the models. The code is available at \url{https://github.com/cm-awais/SAR_sampling}.

\section{Results}
\label{sec:results}
The results section is structured to provide a comprehensive understanding of the findings. It is divided into two main parts: the first focuses on the impact of oversampling across different datasets, analyzing how this technique influences performance metrics. The second explores the effects of oversampling on class imbalances, highlighting its role in adjusting class sizes and improving model performance in scenarios with uneven data distribution.
\subsection{Oversampling and datasets}
The results for FuSARShip (\Cref{tab:fusar_scores}) suggested that the M2m$_u$ enhanced F1-scores, achieving an 8.82\% increase compared to the baseline and even a 0.63\% improvement over the proposed M2m$_f$. While models with ViT and VGG as feature extractors demonstrated improved performance, the model using ResNet as the feature extractor predominantly exhibited signs of underfitting. Notably, the M2m$_u$ method consistently outperformed all other approaches on average, whereas the original M2m$_{orig}$ method yielded the lowest performance among the methods evaluated.

\begin{table}[hbt]
\centering\caption{Performance analysis for FUSARShip.}
\begin{tabular}{cccccc}
\toprule
\multicolumn{1}{c}{\textbf{CLS}}                & \textbf{Models} & \textbf{Baseline} & \textbf{M2m$_{orig}$}      & \textbf{M2m$_f$}         & \textbf{M2m$_u$ }        \\ \toprule
\multicolumn{1}{c}{\multirow{3}{*}{2}} & ViT    & 55.92    & 68.08          & 69.37 & \textbf{69.95} \\ \cline{2-6} 
\multicolumn{1}{c}{}                   & VGG    & 57.23    & 68.7  & 69.37          & \textbf{70.81} \\ \cline{2-6} 
\multicolumn{1}{c}{}                   & ResNet & 55.4     & \textbf{60.23} & 55.4           & 55.4           \\ \toprule
\multicolumn{1}{c}{\multirow{3}{*}{4}} & ViT    & 42.61    & 53.23          & \textbf{67.88} & 61.79          \\ \cline{2-6} 
\multicolumn{1}{c}{}                   & VGG    & 60.91    & 62.72          & 63.3           & \textbf{69.54} \\ \cline{2-6} 
\multicolumn{1}{c}{}                   & ResNet & 48.39    & 12.3           & 42.61          & 42.61          \\ \toprule
\multicolumn{1}{c}{\multirow{3}{*}{9}} & ViT    & 36.95    & 0.92           & 58.52 & \textbf{59.92} \\ \cline{2-6} 
\multicolumn{1}{c}{}                   & VGG    & 53.56    & 45.05          & 60.26 & \textbf{62.38} \\ \cline{2-6} 
\multicolumn{1}{c}{}                   & ResNet & 38.27    & 5.69           & 36.3           & 36.3           \\ \toprule
\multicolumn{2}{c}{AVG}                         & 49.92    & 41.88          & 58.11          & \textbf{58.74} \\ \bottomrule
\end{tabular}

\label{tab:fusar_scores}
\end{table}

In contrast, the results for OpenSARShip (\Cref{tab:opensar_scores}) provided a different trend compared to FuSARShip results. Particularly, M2m$_f$ mostly performed better, achieving an average improvement of F1 of 5.4\% over baseline and 0.98\% compared to the proposed M2m$_u$. Similar to FuSARShip, the models using ViT and VGG as feature extractors improved classification performance, whereas models with ResNet as the feature extractor again showed signs of under-fitting. On average, M2m$_f$ achieved the best performance, while M2m$_{orig}$ emerged as the poorest-performing method.

\begin{table}[hbt]
\centering\caption{Performance analysis for OpenSARShip.}
\begin{tabular}{cccccc}
\toprule
\multicolumn{1}{c}{\textbf{CLS}}                & \textbf{Models} & \textbf{Baseline} & \textbf{M2m$_{orig}$}      & \textbf{M2m$_f$}         & \textbf{M2m$_u$}         \\ \toprule
\multicolumn{1}{c}{\multirow{3}{*}{2}} & ViT    & 63.51    & 63.51          & \textbf{74.1}  & 74.06          \\ \cline{2-6} 
\multicolumn{1}{c}{}                   & VGG    & 67.04    & \textbf{73.16} & 73.07          & 69.57          \\ \cline{2-6} 
\multicolumn{1}{c}{}                   & ResNet & 63.51    & 61.56          & 63.51          & 63.51          \\ \toprule
\multicolumn{1}{c}{\multirow{3}{*}{4}} & ViT    & 59.64    & 15.23          & \textbf{69.36} & 68.46          \\ \cline{2-6} 
\multicolumn{1}{c}{}                   & VGG    & 63.23    & 64.44          & 65.78          & \textbf{68.13} \\ \cline{2-6} 
\multicolumn{1}{c}{}                   & ResNet & 59.63    & 58.5           & 59.64          & 59.64          \\ \toprule
\multicolumn{1}{c}{\multirow{3}{*}{9}} & ViT    & 58.13    & 40.84          & \textbf{70.39} & 68.38          \\ \cline{2-6} 
\multicolumn{1}{c}{}                   & VGG    & 59.94    & 60.23          & \textbf{67.55} & 62.88          \\ \cline{2-6} 
\multicolumn{1}{c}{}                   & ResNet & 57.96    & 38.22          & 57.96          & 57.96          \\ \toprule
\multicolumn{2}{c}{AVG}                         & 61.4     & 52.85          & \textbf{66.82}          & 65.84          \\ \bottomrule
\end{tabular}
\label{tab:opensar_scores}
\end{table}

\subsection{Oversampling and Number of Classes}
\begin{figure}[ht]
    \centering
    \begin{subfigure}[b]{0.472\textwidth}
        \centering
        \includegraphics[width=\textwidth]{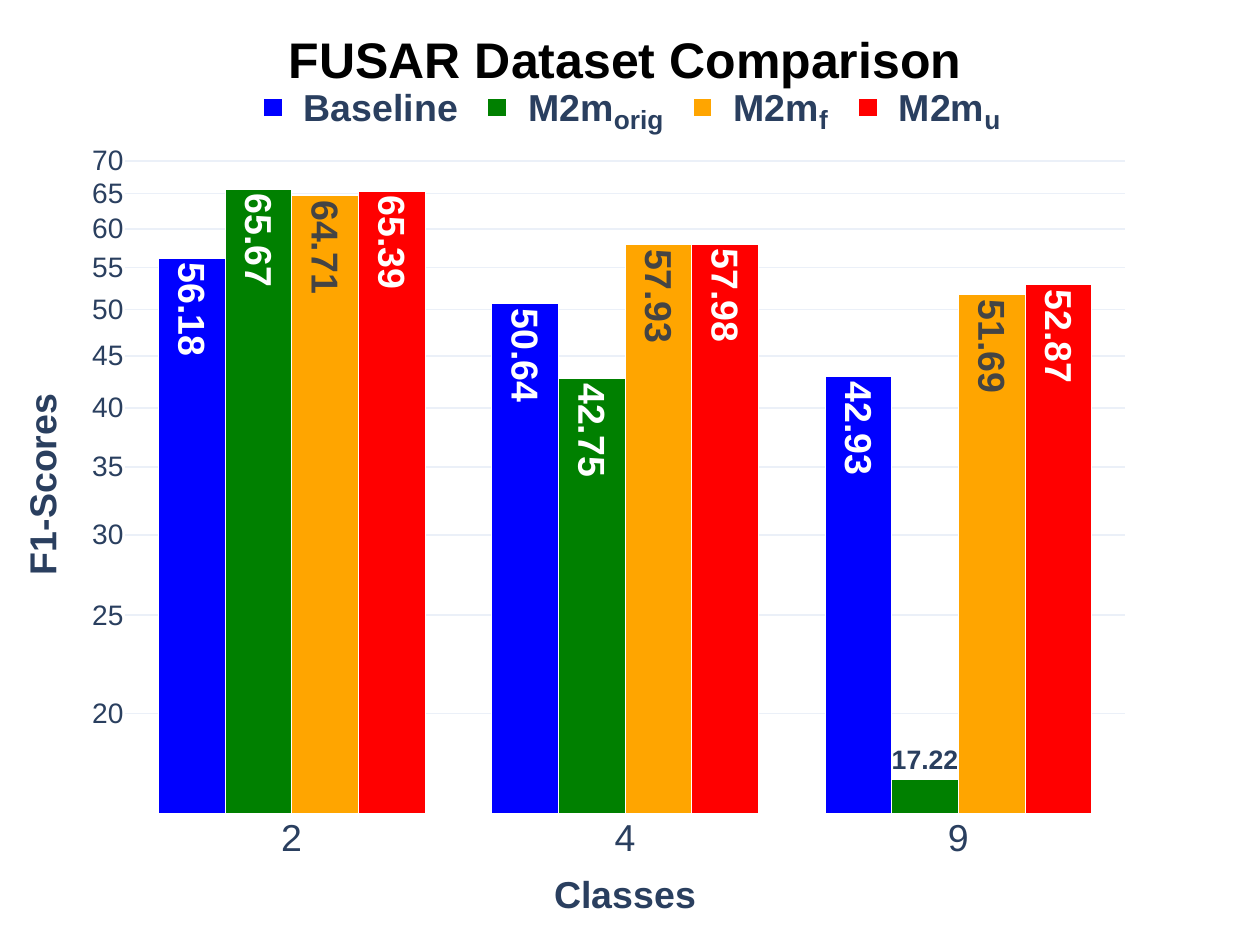}
        \caption{Class-based performance for FuSARShip}
        \label{fig:fusar_f1}
    \end{subfigure}
    \hfill
    \begin{subfigure}[b]{0.472\textwidth}
        \centering
        \includegraphics[width=\textwidth]{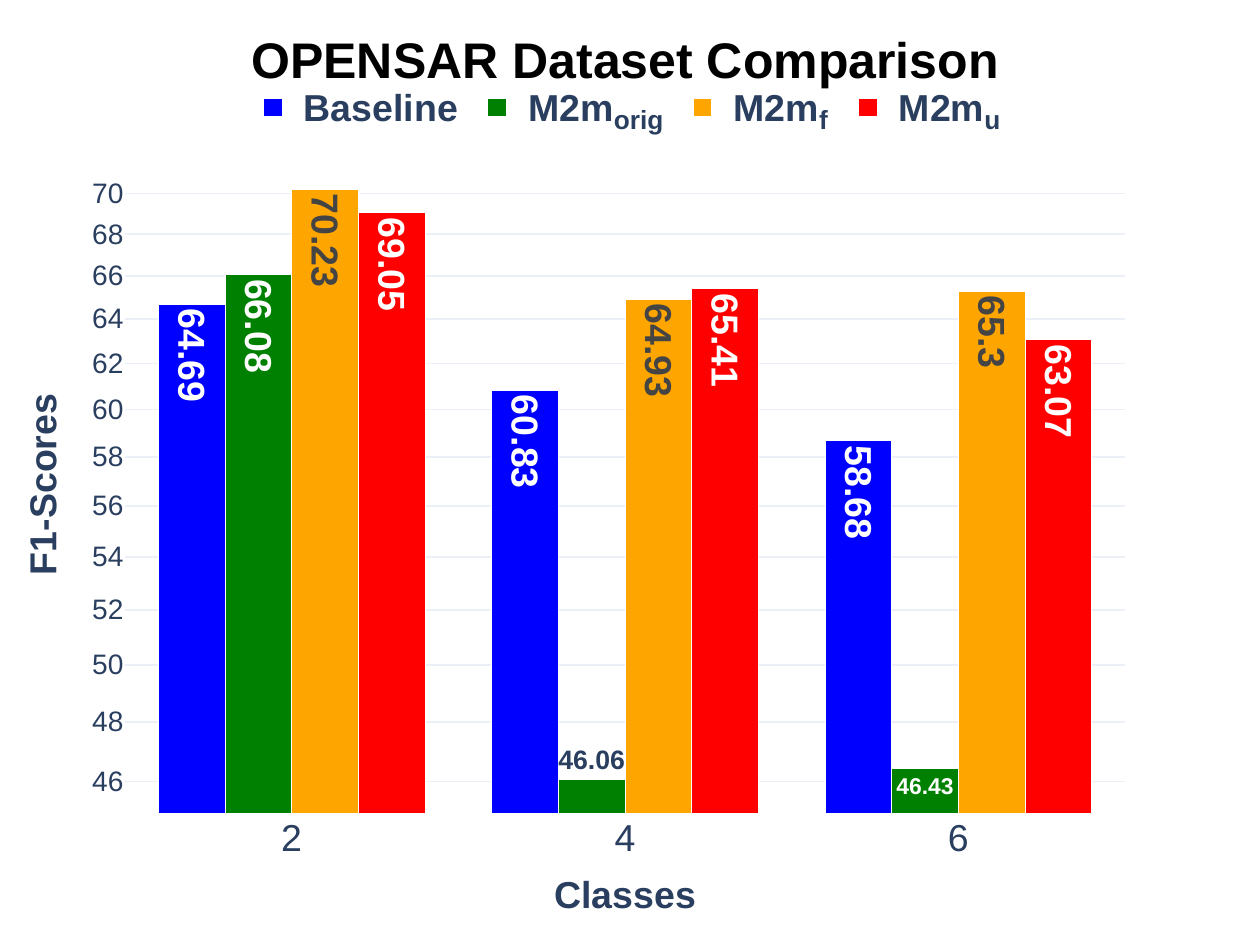}
        \caption{Class-based performance for OpenSARShip}
        \label{fig:opensar_f1}
    \end{subfigure}
    \caption{Class-based F1-score performance comparison for FuSARShip and OpenSARShip datasets, with scores presented on a logarithmic scale for improved visualization.}
    \label{fig:combined_f1_scores}
\end{figure}

Another aspect of our methodology examined the impact of oversampling on different class configurations. For FuSARShip (\Cref{fig:fusar_f1}), M2m$_u$ demonstrated superior performance across 4- and 9-class datasets, while M2m$_{orig}$ excelled in 2-class classification. Similarly, for OpenSARShip (\Cref{fig:opensar_f1}), M2m$_f$ outperformed other methods in 2- and 6-class configurations, whereas M2m$_u$ achieved the best results for 4-class classification. Notably, M2m$_{orig}$ consistently underperformed compared to baselines when applied to datasets with 4 or more classes in both cases.

\section{Discussion}
\label{sec:discussion}
\subsection{Dataset Performance Analysis}
The results for FuSARShip demonstrated the superiority of M2m$_u$, with feature-level methods outperforming both the baseline and the M2m$_{orig}$. This highlights the importance of feature-based oversampling for the FuSARShip dataset. In contrast, the OpenSARShip dataset presented a different trend, with M2m$_f$ outperforming other training strategies. We believe this discrepancy in performance is due to differences in data quality between the two datasets. FuSARShip, being a high-resolution dataset compared to OpenSARShip, exhibited notable differences in the number of synthetic samples added during augmentation. Specifically, 500 synthetic samples were generated for 4- and 9-class configurations in FuSARShip, whereas only 250 synthetic samples were added for 4- and 6-class configurations in OpenSARShip. These disparities in resolution and augmentation scale may have constrained the performance improvements observed in OpenSARShip.

Despite these differences, both methodologies successfully improved performance, achieving an average F1-score increase of 8\% for FuSARShip and 5\% for OpenSARShip, on the other hand, M2m$_{orig}$ at the sample level decreased performance, highlighting the effectiveness of the feature-level oversampling methods.

\subsection{Model Performance Analysis}
No single model emerged as conclusively superior across all scenarios. VGG showed promising results in FuSARShip with an average F1-score of 65.94\%, while ViT achieved 64.57\%. Conversely, in OpenSARShip, ViT outperformed VGG with a 3\% higher F1-score. It is noteworthy that ViT performed worse than VGG in baseline models but achieved competitive results with oversampling interventions, indicating the reliability of ViT when paired with feature-level oversampling methods.

In contrast, ResNet underperformed in all scenarios. We hypothesize that its deeper architecture may have led to underfitting or challenges in handling the limited dataset, while the shallower architectures of ViT and VGG demonstrated greater effectiveness.

\subsection{Class-Based Performance}
We evaluated three different versions of both datasets to analyze the general effects of feature oversampling. For the 4-class versions of both datasets, M2m$_u$ outperformed all other methods. For the 9-class version of FuSARShip, M2m$_u$ also yielded the best results, whereas for the 2-class version of FuSARShip, M2m$_{orig}$ gained the best F1-score. However, for classes 2 and 6 in OpenSARShip, M2m$_f$ significantly outperformed M2m$_u$, achieving up to 2\% better F1-scores. Except for 2 classes of both datasets, M2m$_{orig}$ performed poorer even compared to baselines, indicating the inability of sample-based oversamples for SAR Ship classification datasets.

\subsection{Cost-Effectiveness and Future Recommendations}
Feature oversampling techniques are cost-effective because they eliminate the need to reiterate the feature extraction process across the entire architecture. Once synthetic features are generated, only the classifier requires training. This approach reduces both training time and computational resource usage.

For future work, it is important to note that the parameter values used in the proposed algorithms (M2m$_u$, M2m$_f$) were derived from the original paper \cite{kim2020m2m} and the authors' experience. These values should be tailored to the specific research problem being addressed. This study underscores the effectiveness of feature-based oversampling in SAR ship classification from three perspectives: 
\begin{itemize} 
    \item Improved performance on imbalanced datasets. 
    \item Cost-effective training and resource utilization. 
    \item Simplicity and adaptability to diverse imbalanced datasets.
\end{itemize}

The results not only highlight the effectiveness of feature space oversampling for SAR ship classification datasets, but also leave room for future research on sample space oversampling techniques dedicated to SAR data, or the development of more robust feature space oversampling techniques.

\section{Conclusion}
\label{sec:conclusion}
In this study, we proposed two feature-level oversampling algorithms that significantly enhanced SAR ship classification. The robustness of our proposed strategies proved to be applicable to SAR datasets, and have the potential to be extended to other fields of research. The results also indicated the lack of performance for sample space oversampling methods for SAR datasets. The proposed method demonstrated a notable increase in classification performance across different imbalanced classes and datasets, highlighting the importance of feature-level approaches in SAR ship classification.

In the future, our goal is to develop SAR-dedicated sample space oversampling methods, and refine feature space oversampling further and extend its application to additional classes and datasets, paving the way for more robust solutions in SAR ship classification and beyond.

\newpage
\bibliographystyle{IEEEtranN}
\bibliography{references}

\end{document}